\documentclass[]{article}
\usepackage{ijcai16}

\usepackage{times,amssymb}
\usepackage[super]{nth}
\usepackage{graphicx,amsmath,url}
\usepackage[ruled,vlined,linesnumbered]{algorithm2e}
\usepackage{multirow}
\usepackage{graphicx} \usepackage{multirow}
\usepackage{caption}
\usepackage{subcaption}

  \newtheorem{definition}{Definition}
  \newtheorem{proposition}{Proposition}

\newcommand{\tn}[1]{\textnormal{#1}}

\newcommand{\vl}{VirtualLabs}

\DeclareMathOperator*{\argmax}{\arg\!\max}
\usepackage{color-edits}

\newcounter{algoline}

\newcount\Comments  % 0 suppresses notes to selves in text
\definecolor{darkgreen}{rgb}{0,0.6,0}
\definecolor{orange}{rgb}{1,0.4,0}
\newcommand{\kibitz}[2]{\ifnum\Comments=1{\textcolor{#1}{#2}}\fi}

\begin{document}
\title{Sequential Plan Recognition}
\author{
 Reuth Mirsky \and Roni Stern \and Ya'akov (Kobi) Gal \and Meir Kalech\\
        Department of Information Systems Engineering \\ Ben-Gurion
  University of the Negev, Israel \\ 
  \{dekelr,sternron,kobig,kalech\}@bgu.ac.il}
 
\maketitle
	
\begin{abstract}
  Plan recognition algorithms infer agents' plans from their observed
  actions. Due to imperfect knowledge about the agent's behavior and
  the environment, it is often the case that there are multiple
  hypotheses about an  agent's plans that are consistent with the
  observations, though only one of these hypotheses is correct.  This
  paper addresses the problem of how to disambiguate between
  hypotheses, by querying the acting agent about whether a candidate
  plan in one of the hypotheses matches its intentions. This process is
  performed sequentially and used to update the set of possible
  hypotheses during the recognition process.  The paper defines the
  sequential plan recognition process (SPRP), which seeks to reduce the number of hypotheses
  using a minimal number of queries.
  We propose a number of  policies  for the SPRP which use maximum likelihood and information gain to choose which plan to query. We show this approach works well in practice on two domains from the literature, significantly reducing the number of hypotheses  using  fewer queries than  a baseline approach.
  Our  results  can inform the design of future plan recognition systems that interleave the recognition  process with intelligent interventions of their users.
\end{abstract}

\section{Introduction}

Plan recognition (PR), the task of inferring agents' plans based on their
observed actions, is a fundamental problem in AI, with a broad range
of applications, such as inferring transportation
routines~\cite{liaoetal07}, advising in health
care~\cite{Allen2006500}, or recognizing activities in gaming and
educational software~\cite{uzan13}.

%The focus of this paper is on recognizing an agent's complete plan hierarchy in real-time, as the agent is performing its activities.

A chief problem facing PR algorithms is how to disambiguate
between multiple  hypotheses (explanations) that are consistent with an
observed agent's activities.
 A straightforward solution to
this problem is to ask the observed agent to reveal the correct
hypothesis, but soliciting complete hierarchies is time and
information consuming and prone to error. Alternatively, querying
whether a given hypothesis is correct will not contribute any
information about the correct hypothesis should the answer be
``false''.
 Consider for example an e-learning
software for chemistry laboratory
experiments~\cite{gal2015making,yaron2010chemcollective}.
There are many possible solution strategies that students can use to solve problems, and variations within each due to exploratory activities and mistakes carried out by the student.
Given a set of actions performed by the student, one hypothesis may relate a given action to the solution of the problem, while another may relate this action to a failed attempt or a mistake. The space of possible hypotheses can become very large, even for a small number of observations.
To illustrate,
% six observations produced (on average) over 4,000
in the chemistry domain
%while seven
just seven observations produced over 11,000 hypotheses on average, with some instances
producing over 32,000 hypotheses.

However, in many domains it is possible to query (for a cost) the observed agent itself or a
domain expert about certain aspects of the correct hypothesis~\cite{kamar2013modeling}. For example, the student may be asked questions about her solution strategy for a chemistry problem during her interaction
with the educational software.
%or offline after all %observations were
% collected (for example, a system administrator observing suspicious
% behavior that can ask a cyber security expert for an opinion regarding these actions).
Answers for such queries allow to reduce the set of
possible hypotheses without removing the correct hypothesis. (e.g., interrupting students may disrupt their
learning and incur a cognitive overhead).

The first contribution of this paper is to define the {\em sequential plan
  recognition process} (SPRP),  in which we
iteratively query whether a given part in one of the hypotheses is
correct, and update all hypotheses in which this plan appears (or
does not appear, depending on the answer to the query).
 We represent a hypothesis as a set of plans, one for each goal that
 the agent is pursuing. This allows to capture settings in which
 agents may pursue several goals at the same time and in which their
 actions may include mistakes (e.g., students performing exploratory
 activities in the lab and trial-and-error).

A key challenge in the SPRP is how to update the hypothesis space following the results of a query. Because recognition is performed in real-time, the hypothesis set may contain incomplete plans that describe  only the observations seen thus far. Thus, for example, if the result of a query on a plan $p$ is true (i.e., the agent plans to perform  $p$), we cannot simply discard all hypotheses that do not contain $p$, because they may contain plans that will evolve to perform $p$ in the future. To address this challenge we developed  criteria for
determining whether possibly incomplete plans appear in the
correct hypothesis. We show that SPRP using these criteria is both sound and complete.

The second contribution of this paper is how to compute a good policy for the SPRP for  choosing which plan in the current set of hypotheses to query.  We consider queries that maximize the information-gain and the
likelihood of the resulting hypotheses given the expected query result.
The third contribution of this paper is to  evaluate
approaches for solving the sequential plan recognition problem in two
domains from the plan recognition literature that exhibit varying degrees of
ambiguity. 
One  of the domains was synthetically
generated~\cite{kabanza2013controlling}, % by configuring the branching
% factor, depth and ordering constraints of generated plans.
while the other logs were taken from real students' traces when
interacting with the aforementioned virtual chemistry
 lab~\cite{AG13,amirAndGal2010}.
 %We  considered candidate plans to query
% and selected the one that maximizes the information gain as well as the likelihood of the
% resulting hypotheses given the expected query result.
%two approaches   from sequential diagnosis for  choosing which candidate plan to query,
In both domains, our approach significantly decreased the number
of queries compared to a baseline technique. The number of queries
performed by the information-gain approach was significantly smaller
than the alternative approaches.
 
\section{Related Work}
Our work relates to different approaches in the PR literature on
disambiguation of the hypothesis space during run-time.
Most of the approaches admit all of the hypotheses that are consistent
with the observed history and rank them~\cite{GeibGoldman09,wiseman2014discriminatively}.

Few works exist on interacting with the observed agent as means to
disambiguate the hypothesis space during plan recognition:  Bisson and Kabanza~\shortcite{bisson2011provoking}
who ``nudge'' the agent to perform an action that will disambiguate
between two possible goals.  Fagundes et
al.~\shortcite{fagundes2014dealing} make a decision to query the
observed agent if the expected time to disambiguate the hypothesis
space does not exceed a predefined deadline to act in response to the
recognized plan. They ask the observed agent directly about its
intentions and do not prune the hypothesis space.  They evaluate their approach in a simulated domain.
We solve an orthogonal problem in
which the observed agent can be queried repeatedly, and the hypothesis
space is pruned based on the query response. We consider the cost of
this query and evaluate the approach in a real-world domain.

Lastly, the deployment of probes, tests, and sensors to identify the correct diagnoses or the occurrence of events was inspired
by work  in sequential diagnosis~\cite{feldman2010model,siddiqi2011sequential},  active diagnosis~\cite{sampath1998active,haar2013optimal}, and sensor minimization~\cite{cassez2008fault,debouk2002optimization}.
~\cite{keren2014goal} suggested a metric that will allow an agent to recognize plans earlier.
%These works don't consider incomplete plans or that
%the hypothesis space dynamically changes over time.
 
\section {Background}
\label{sec:probDef}
Before defining the SPRP we present some background about plans and PR.
There are multiple ways to define a plan and the PR
problem~\cite[inter alia]{Nau:2007tf,ramirez2010probabilistic}. We follow the definitions used by Kabanaza et
al.~\shortcite{kabanza2013controlling} (simplified for brevity) in which  the
observing agent is given a {\em plan library} describing the expected
behaviors of the observed agent.
\begin{definition} (Plan Library)
%A plan library is a tuple $L=\langle B,C,R \rangle$, where $B$ is a finite set of basic actions, $C$ is a finite set of complex actions, and $R$ is a set of refinement methods
A plan library is a tuple $L=\langle B,C,R \rangle$, where $B$ is a set of basic actions, $C$ is a set of complex actions, and $R$ is a set of refinement methods
of the form $c\rightarrow (\tau,O)$, where (1) $c\in C$; (2) $\tau \in (B \cup C)^*$; (3) and $O$ is a partial order over $\tau$ representing ordering constraints over
the actions in $\tau$.
\label{def:plan-library}
\end{definition}
The refinement methods represent how complex actions can be decomposed into (basic or complex) actions.
A {\em plan} for achieving a complex action $c\in C$ is a tree whose root is labeled by $c$, and each parent node is labeled with a complex action such that its children nodes are a decomposition of its complex action into constituent actions according to one of the refinement methods.
The ordering constraints of each refinement method are used to enforce the order in which the method's constituents were executed ~\cite{GeibGoldman09}.

 A plan is a labeled tree $p=(V,E,\mathcal{L})$, where $V$ and $E$ are the nodes and edges of the tree, respectively,
 %$r$ is the root of the tree,
 and $\mathcal{L}$ is a labeling function $\mathcal{L}: V \rightarrow B\cup C$ mapping every node in the tree to either a basic or a complex action in the plan library. Each inner node is labeled with a complex action such that its children nodes are a decomposition of its complex action into constituent actions according to one of the refinement methods.

The set of all leaves of a plan $p$ is denoted by $leaves(p)$, and a plan is said to be {\em complete} iff
all its leaf nodes are labeled basic actions, i.e., $\forall v\in leaves(p), \mathcal{L}(v)\in B$.
%A plan is said to be {\em complete} if all its leaf nodes are labeled with basic actions.

 An \emph{observation sequence} is an  ordered set of basic actions that represents actions carried out by the observed agent.
 A plan $p$ \emph{describes} an observation sequence $O$ iff every observation is mapped to a leaf in the tree. More formally, there exists an injective function $f:O\rightarrow leaves(p)\cap B$ such that $f(o)=v$.
 %i.e., every observation in $O$ is mapped to a basic action in $T$.
 The observed agent is assumed to plan by choosing a subset of complex actions as intended goals and then carrying out a separate plan for completing each of these goals.

Importantly, an agent may pursue several goals at the same time. %, especially during exploratory activities.
Therefore, a hypothesis can include a set of plans, as described in the following definition:
\begin{definition} (Hypothesis)
\label{def:hyp}
 A hypothesis for an observation sequence
  is a set of plans such that each plan describes a mutually
  exclusive subset of the observation sequence and taken together the   plans describe all of the observations.  We then say that the hypothesis describes the observation sequence.
\end{definition}

To illustrate these concepts we will use a running example from an
open-ended educational software package for chemistry called {\vl},
which also comprises part of our empirical analysis. {\vl} allows
students to design and carry out their own experiments for
investigating chemical processes~\cite{yaron2010chemcollective} by
simulating the conditions and effects that characterize scientific
inquiry in the physical laboratory.  

Such software is open-ended and
flexible and is generally used in classes too large for teachers to
monitor all students and provide assistance when needed. Thus, there
is a need to develop recognition tools to support teachers'
understanding of students' activities using the software.

We use the
following problem  as a running example:
Given four substances $A;B;C$, and $D$ that react
  in a way that is unknown, design and perform virtual lab experiments
  to determine which of these substances react, including their
  stochiometric coefficients.

\begin{figure}[t]
 \centering
 \includegraphics[width=8cm]{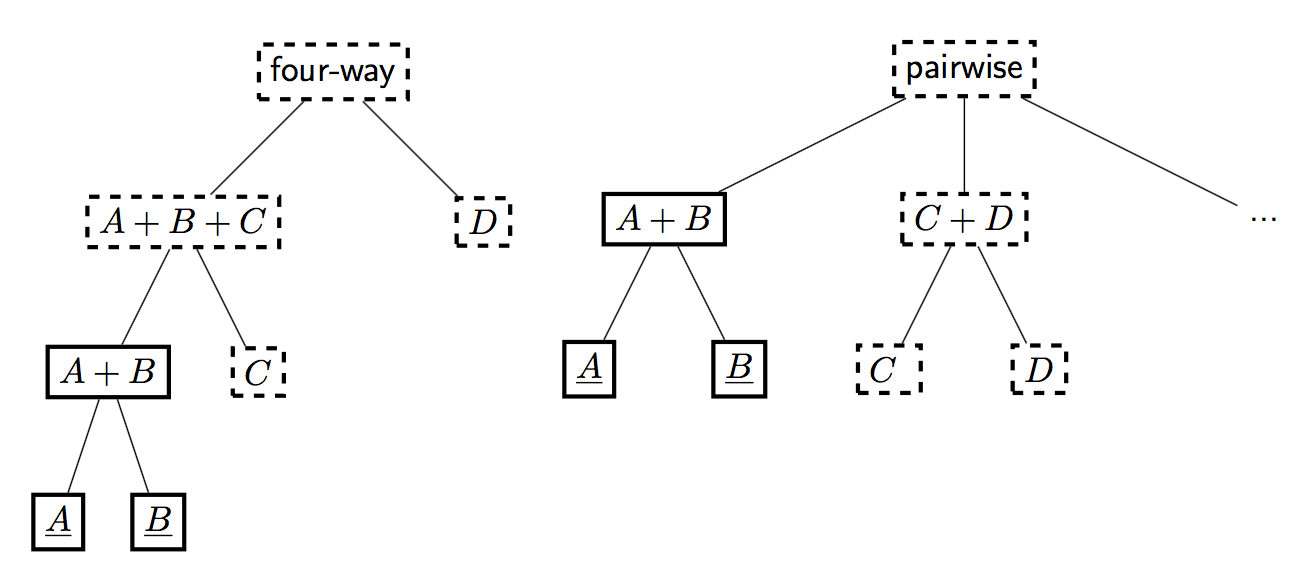}
 \caption{Two candidate hypotheses in {\vl} for observations $A$ and $B$.}
 \label{fig:example}
\end{figure}

There are two classes of strategies used by students to solve the
above problem in {\vl}.  The most common strategy, called
\emph{pairwise}, is to mix pairs of solutions ($A$ with $B$, $A$ with
$C$, etc.) in order to determine which solutions react with one
another. % In some cases, students stop mixing pairs when a reaction
% occurs because this is sufficient to identify the reactants.
In the \emph{four-way} solution strategy, all substances are mixed in
a single flask, which is sufficient to identify which solution pair were
the reactants and which did not react, since the non-reactant are still observable after the reaction.

Now suppose that the student is observed to mix solutions $A$ and $B$
together in a single flask. Without receiving additional information,
both the pairwise and four-way strategies are hypotheses that are consistent
with the observations, and both include an \emph{incomplete plan} describing the student's actions.

Incomplete plans include nodes labeled with complex level actions
that have not been decomposed using a refinement method. These
\emph{open frontier} nodes represent activities that   the agent will carry out in future and have yet to be refined.  
This is similar   to
the least commitment policies used by some planning approaches to
delay variable bindings and commitments as much as
possible~\cite{Tsuneto:1996wl,AvrahamiKaminka05,avrahami2007incorporating}.

This ambiguity is exemplified in
Figure~\ref{fig:example}, showing one hypothesis for the four-way
solution strategy (left) and one for the pairwise solution strategy
(right).  Each of these hypotheses contain a single incomplete plan. The nodes representing the observations $A$ and $B$ are
underlined.  The dashed nodes denote open frontier nodes.

We can now define the plan recognition problem.
\begin{definition} (Plan Recognition (PR))
\label{def:pr}
	A PR problem is defined by the tuple $\langle L, O \rangle$
	where $L$ is a plan library and $O$ is an observation seqeuence.
	A PR algorithm accepts a PR problem and
	outputs a set of hypotheses $H$ such that each hypothesis describes
	the observation sequence.
  \end{definition}

	Let $h^*$ be the \emph{correct}
  	hypothesis, i.e., the set of  plans the agent intends to
  	follow ($h^*$ is not known at recognition time). When recognition is performed in real-time,  observations are collected over time, there is uncertainty about future activities,  and the agent's plans may be incomplete (e.g., the agent may have not decided how to perform some of the planned complex actions).  To address this challenge we require the following notion of {\em plan refinement}.

\sloppy{
\begin{definition} (Refinement of a  plan)
  A plan $p$ is a refinement of a plan $p'$, denoted by $p'\sim^r p$,
  if the plan $p$ can be obtained by applying a (possibly empty) sequence of refinement
  methods from the plan library $L$ to $p'$.
\label{def:refinement}
\end{definition}}
The refinement criterion is asymmetric and transitive. Note that a plan can always refined from itself using an empty sequence of refinement methods. We  extend the refinement criteria to hypotheses as follows.
A hypothesis $h$ is a refinement of a hypothesis $h'$, denoted ($h' \sim^r h$), if there is a one-to-one mapping between every plan $p\in h $ and a plan $p'\in h'$ such that $p$ is a refinement of $p'$ ($p'\sim^r p)$.

Using this definition, a PR algorithm is complete if it returns a hypothesis set $H$ such that $h \sim^r h^* \rightarrow h \in H$, that is, H contains all possible hypotheses that can be refined to $h^*$.% is a refinement of some hypothesis $h\in H$.

\begin{figure}
\centering
% \begin{subfigure}{.60\textwidth}
%   \centering

\includegraphics[width=8cm]{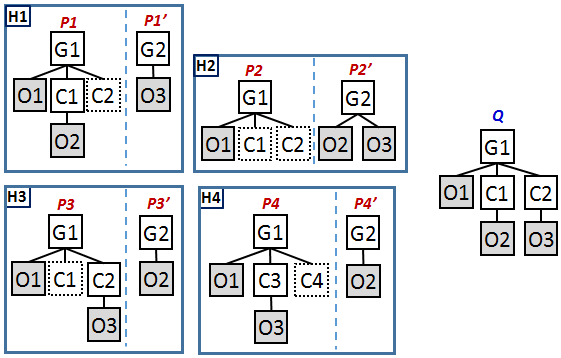}
  \caption{Four candidate hypotheses $H1$, $H2$, $H3$, and $H4$ for observations $O1$, $O2$, and $O3$, and one complete plan $Q$.
} %and four complete plans $Q1$,    $Q2$, $Q3$, and $Q4$ that each describes a subset of the observations. The green arrows shows which complete plans are refinement of the plans in they hypotheses.}
  \label{fig:Exp1}
\end{figure}

To illustrate, the top part of Figure~\ref{fig:Exp1} shows part of a  hypothesis set
$H1,\ldots,H4$, each of these hypotheses explains the observations $O1$,
$O2$, and $O3$.
The figure  shows the four possible hypotheses for the
observation sequence.  Each hypothesis $Hi$ ($i=1,\ldots,4$) consists
of two plans, $P_i$ and $P'_i$. Nodes in gray represent the
observations and nodes with dashed outline represent open-frontier actions.

% An arrow between plans $Pi$ and $Qi$ represents that $Qi$ is a refinement of $Pi$ ($Pi\sim^rQi$). In addition to these relations,
The plans $P1$ and $P3$ are both refinements of $P2$ ($P2\sim^r P1$ and $P2\sim^r P3$), but they are not refinement of each other
($P1\not\sim^r P3$ and $P3\not\sim^r P1$).  In addition, $H1$ is not a
refinement of $H2$ ($H2\not\sim^r H1$) because the plan $P1'$ is not a
refinement of $P2'$ $(P2'\not\sim^r P1')$. Similarly, $H3$ is not a
refinement of $H2$ ($H2\not\sim^r H3$).

\section{Sequential Plan Recognition}

 In this section we define the SPRP, beginning with the notion of a query function and a query policy.
\begin{definition} (Query Function)
\label{def:query}
  A query $QA$ is a function that receives as input a plan $p$ and
  outputs whether one of the plans in the correct
  hypothesis $h^*$ can be refined from $p$.
 \begin{equation}
   \label{eq:1}
  QA(p)=
\begin{cases}
  \tn{True}  & \tn{if } \exists p'\in h^* \tn{ s.t. } p \sim^r p' \\
 \tn{False} & \text {otherwise}
\end{cases}
\end{equation}
\end{definition}

A query policy  selects which plan to query in a SPRP given the current set of hypotheses.
 \begin{definition} (Query Policy)
\label{def:policy}
   A query policy is a function $\pi : \mathcal{H} \rightarrow \mathcal{P}_\mathcal{H}$, where $\mathcal{H}$ is the set of all possible hypotheses and $\mathcal{P}_\mathcal{H}$ is the set of all plans in all possible  hypotheses. %$\mathcal{P}[\mathcal{H}]=\{p \mid p\in h, h\in \mathcal{H}\}$.
   %   from   the space of all possible hypotheses $\mathcal{H}$ to the space of all plans in all possible  hypotheses $P=\{p \mid p\in h, h\in \mathcal{H}\}$.
 \end{definition}
 Such a policy needs to trade off the immediate benefits of a query with the short and long term costs associated with disrupting the acting agent.

 Given an initial hypothesis set $H_0$ (obtained by applying a PR algorithm),
 a query function $QA(\cdot)$, and a query policy $\pi$,  the \emph{Sequential Plan Recognition Process (SPRP)}   is the iterative process shown in
 Algorithm~\ref{alg:sprp}.  Starting from the initial iteration $i=0$, in every iteration of SPRP, a candidate plan $p$ is chosen from the set of hypotheses $H_i$ and the result of the query is used to generate an updated hypothesis set $H_{i+1}$ to be used in the following iteration. We maintain a $CLOSED$ list of
the chosen hypotheses up to step $i$, and terminate when there are no
more plans in the hypothesis set left to query or there is just  a single hypothesis in the set $H_i$
(line 3). The output of the algorithm is the set $H_i$ at the last iteration.

\begin{algorithm}[t]
%\KwIn {$L$ is a plan library,}
%\KwIn {$H_0$ is the set of hypotheses outputted by a plan recognition algorithm}
\KwIn {$H_0$ is the initial set of hypotheses}
\KwIn {$QA$ is a query function}
\KwIn {$\pi$ is a query policy}
%\algoLine
$i\gets 0$; $CLOSED \gets \emptyset$\\
%\While{$\bigcup_{h\in H_i}h\setminus CLOSED$ is not empty, or $|H_i|=1$}
\While{$\bigcup_{h\in H_i}h\setminus CLOSED\neq \emptyset$ or $|H_i|=1$}
{
	$p\gets \pi(H_i)$\\
$H_{i+1}\gets \tn{Update}(QA(p),H_i,p)$
\label{line:update}\\
	$i\gets i+1$\\	
	Add $p$ to $CLOSED$ \\
}
  \caption{Sequential Plan Recognition Process.}
  \label{alg:sprp}
\end{algorithm}

The key step in the algorithm is how $H_i$ should change after
performing a query on a plan $p$ (line 5).  Suppose
$QA(p)=\tn{True}$. According to Definition~\ref{def:query}, this means
that there exists a plan $p^*\in h^*$ such that $p^*$ is a refinement
of $p~(p \sim^r p^*)$.  If we knew $p^*$ we could simply remove from
$H_i$ all hypotheses that do not contain a plan $p'$ that can be refined
to $p^*$ ($p'\sim^r p^*$).

Since we do not know $p^*$, a natural option is to remove from $H_i$ all
the hypotheses that  do not have any plan $p'$ that can be refined from $p$. However, in certain situations this may lead us to discard the
correct hypotheses.

Consider the example of Figure~\ref{fig:Exp1} %in which $QA(P1)=\tn{True}$,
and assume that we query plan $P1$ which returns true (i.e., $QA(P1)=\textnormal{True}$).
If we remove all hypotheses that do not contain plans that are  refinements of $P1$, then
hypothesis $H3$ will be removed, since neither $P3$ nor $P3'$ are refinements of $P1$.
However, it may be the case that one of the agent's intended plans is plan $Q$ (right of Fig.~\ref{fig:Exp1}).
The query on $P1$ returned true because $P1\sim^r Q$. However, note that $H3$ is
a valid hypothesis and should not be discarded, since $P3\sim^r Q$. Thus, we require a different pruning criteria for the hypotheses,
given an outcome of query.

To handle this problem, we need to devise a new criteria for
determining whether two plans can be used to refine a third plan. We
will present this criteria and then show how it can be used to update
the set of hypotheses for the next time step in a way that preserves
the completeness of the PR process.
\begin{definition} (Matching of Plans)
\label{def:match}
A pair of plans $p$ and $p'$ are said to match,
  denoted by $p'\sim^m p$ (or $p \sim^m p'$), if there exists a plan
  $p''$ that is a refinement of both plans $p$ and $p'$ ($p\sim^r p''$ and $p'\sim^r p''$).
\end{definition}
Note that the match criteria is symmetric.
To illustrate this concept using the example in Figure~\ref{fig:Exp1}, the plan $P1$ matches $P3$
($P1 \sim^m P3$), even though they are not refinements of each other,
since there is at least one plan  which is
a refinement of both. The complete plan $Q$ is an example of such a plan, since $P1 \sim^r Q$ and $P3 \sim^r Q$.

Using both the {\em match} (Definition~\ref{def:match}) and {\em
  refinement} (Definition~\ref{def:refinement}) relations, we define the update rule (Algorithm~\ref{alg:sprp}, line 4) over the hypothesis set $H$ which depends on whether the query
$QA(p)$ returns True or False: \\
{\bf Case 1: $QA(p)=\tn{True.}$} For
this case we define the set $\phi(H,p,\tn{True})$
which includes only hypotheses in which at least one of the plans match $p$:
 \begin{equation}
    \label{eq:updateT}
\phi(H,p,\tn{True}) =
  \{ h \mid h\in H \wedge \exists p'\in h ~~ p'\sim^m p \}
 \end{equation}
In our example in Figure~\ref{fig:Exp1}, if $QA(P1)=\tn{True}$ then
we know that the correct hypothesis $h^*$ will contain a complete plan
that is a refinement of $P1$. In particular,  $Q$ is a possible refinement of $P1$.  Thus, any
hypothesis $h\in \{H1\ldots,H4\}$ that has at least one plan $p$ that
can be refined to $Q$ (or any other plan that is a  refinement of $P1$) cannot be pruned. Therefore,
the hypothesis $H2$ is not pruned, because $Q$ is a
refinement of $P2$ ($P2\sim^r Q$). Similarly, the hypothesis $H3$ is
not pruned, because $Q$ is a refinement of $P3$ ($P3\sim^r Q$).
However, the hypothesis $H4$ is pruned since there is no plan in it
that can be refined to a plan that is also a refinement of
$P1$.  %In general, we can safely prune all hypotheses that do not contain a plan that matches the queried plan $p$.

{\bf Case 2:
 $QA(p)=\tn{False.}$}
This means that there is no plan $p^*\in h^*$ that is a refinement of $p$.
The refinement operator is transitive, i.e., if $p''$ is a refinement of $p'$ and $p'$ is a refinement of $p$, then $p''$ is also a refinement of $p$.
Therefore, if $h^*$ does not contain any plan that is a refinement of $p$, we can safely remove from $H$ every hypothesis that contains a plan $p'$
such that $p'$ is a refinement of $p$.
\begin{equation}
    \label{eq:updateF}
\phi(H,p,\tn{False}) =
   H \setminus \{ h \mid h\in H \wedge \exists p'\in h ~~ p\sim^r p' \}
%h \mid h\in H_i \wedge \exists p'\in h ~~ p'\sim^m p \}
 \end{equation}
%Therefore, the   set of hypotheses $H_{i+1}$ is updated as follows:
% \begin{equation}
%    \label{eq:updateF}
% H_{i+1}=
% \{ h \mid h\in H_i \wedge \neg\exists p'\in h ~~ p\sim^r p' \}
% \end{equation}
In our example in Figure~\ref{fig:Exp1}, if $QA(P2)=\tn{False}$,
 there does not exist any plan in $h^*$ that is a refinement of
$P2$. Therefore, we can safely remove hypotheses $H1$, $H2$, and $H3$, because each of them
has at least one plan that is a refinement of $P2$ (formally, $P2\sim^r P1, P2\sim^r P3$, and  $P1\sim^r P1$).
If  $QA(P1)=\tn{False}$, then  only $H1$ is pruned.

We can now  define the  update rule (line 4) for the Sequential Plan Recognition Process  as follows:
\begin{multline*}
\noindent
%   \label{eq:4}
   \small
\tn{Update}(QA(p),H_i, p)=
 \begin{cases}
\phi(H_i,p,\tn{True})  &  $QA(p)=\tn{True.}$ \\
\phi(H_i,p,\tn{False})  &   \tn{ otherwise }
 \end{cases}
 \normalsize
 \end{multline*}
 
%We assume that the PR process is complete, meaning that if it outputs a set $H_0$ such that $\forall h ~~ h \sim^r h^*$, it holds that $h \in H_0$. Using this assumption,
We assume that the PR algorithm is complete and provides a set of probability-ranked hypotheses, as is common in the state-of-the art. We can  now state that SPRP described in Algorithm~\ref{alg:sprp} is both sound and complete:

\begin{proposition}
The SPRP will necessarily terminate in a finite number of iterations
$k$ with a hypothesis set $H_k \subseteq H_0$ such that  the following holds:
\begin {description}
%\item [Completeness] $\forall h\in H_0, ~~ h \sim^r h^* \rightarrow h \in H_k$. SPRP does not remove any hypothesis that can be refined to the correct hypothesis $h^*$.
%\item [Soundness] $\forall h \in H_k, ~~ h \sim^r h^*$. Every hypothesis SPRP keeps can be refined to the correct hypothesis $h^*$
\item [Completeness] SPRP does not remove any hypothesis that can be refined to the correct hypothesis $h^*$. Formally, $\forall h\in H_0, ~~ h \sim^r h^* \rightarrow h \in H_k$. 
\item [Soundness] Every hypothesis SPRP keeps can be refined to the correct hypothesis $h^*$. Formally, $\forall h \in H_k, ~~ h \sim^r h^*$. 
\end{description}
\label{prop:complete}
\end{proposition}

\noindent {\bf Termination} First, we must show that after a finite number of iterations, the SPRP will terminate. This is  immediate, since at each iteration we ask about a plan from the remaining set of plans. This means that at the worst case, if no hypothesis is removed, the process will terminate after $\mid T \mid$ iterations, where $T$ is the set of all  plans in all  hypotheses.

\noindent {\bf Completeness}
%The proof  depends on showing that at each step $i$ in the algorithm,  we do not discard any $h$ such that $h^*$ can be refined from $h$. Formally, $\forall h \in H_0 ~~ (h \sim^r h^*) \rightarrow h \in H_i$. 
We prove completeness by showing that every $h$ that was removed from $H_0$, could not be refined to $h^*$.
This reasoning follows  from the update rule in each case of examining some plan $p$:
\noindent If  $QA(p)= \tn{True}$, then
\begin{multline*}
QA(p)=\tn{True} \Rightarrow  \exists p^*\in h^* ~~ p\sim^r p^*  \\
 \Rightarrow \forall h\in H ~~ h\sim^r h^* \rightarrow \exists p'\in h ~~ p'\sim^r p^* 
% \Rightarrow \forall h\in H ~~ h\sim^r h^* \rightarrow  \exists p'\in h ~ \exists p'' ~~ p'\sim^r p'' \wedge p\sim^r p'' \\
% \Rightarrow p'\sim^r p^* \wedge p\sim^r p^* \rightarrow p \sim^m p' \\
% \Rightarrow \forall h\in H ~~ \neg \exists p'\in h ~ p\sim^m p'  \rightarrow \neg (h\sim^r h^*)
\end{multline*}
We can conclude that if $\forall p'\in h$ do not match the query plan $p$, we can safely remove the hypothesis $h$ because $h^*$ cannot be refined from  $h$.
\noindent If  $QA(p)= \tn{False}$, then the following holds:
\begin{multline*}
QA(p)= \tn{False} \Rightarrow  \forall p^* \in h^* ~~ \neg (p \sim^r p^*) \\ %\neg\exists p^*\in h^* ~~ p\sim^r p^*  \\
%\Rightarrow  \forall p^*\in h^* ~~ \neg(p\sim^r p^*)  \\
% \Rightarrow \forall h\in H ~~ h\sim^r h^* \rightarrow \exists p'\in h ~~ p'\sim^r p^* \\
% \Rightarrow \forall h\in H ~~ h\sim^r h^* \rightarrow  \neg\exists p'\in h ~~ p'\sim^r p \\
%\Rightarrow \forall h\in H ~ \exists p'\in h ~~ p'\sim^r p \rightarrow \neg (h\sim^r h^*)
\Rightarrow \forall h \in H ~~ \exists p' \in h ~~ p' \sim^r p \rightarrow \neg (p' \sim^r p^*) \\
\Rightarrow \forall h \in H ~~ \exists p' \in h ~~ p' \sim^r p \rightarrow \neg (h \sim^r h^*) 
\end{multline*}
Thus, we can conclude that if the query plan $p$ can be refined from  $p'\in h$, we can safely remove the hypothesis $h$ because $h^*$ cannot be refined from  $h$.

\noindent {\bf Soundness}
Let $H_k$ be the set of all hypotheses after $k$ iterations and $h^*$ is the correct hypothesis.
If there is still a hypothesis $h \in H_k$ such that $\neg(h \sim^r h^*)$, then $\exists p \in h ~ ~ \forall p^* \in h^* \neg(p \sim^r p^*)$.
Thus, we can still query about $p$ and $k$ is not the final iteration of the algorithm. Hence, at the final iteration of the algorithm we have that $\forall h ~~ h \sim^r h^*$.

\section{Probing Techniques}
%  Finding the optimal query policy can be done by formalizing the SPRP as a planning-under uncertainty problem.  The set of actions is which plan to select, and the state-space in this formalization needs to include all possible sets of hypotheses, which is infeasible. 
 We   propose several heuristic methods for generating a PR policy that aim to minimize the number of queries required to achieve the minimal set of hypotheses that are consistent with the observation. These methods rely on the standard assumption that each hypothesis $h$ is associated with a 
 lity $P(h)$ that is assigned by the PR algorithm  (such as PHATT, DOPLAR and ELEXIR~\cite{GeibGoldman09,kabanza2013controlling,geib2009delaying}).

\noindent {\bf \emph{Most Probable Hypothesis (MPH)}.} Choose a plan from the hypothesis $h$ that is associated with the highest probability and was not yet queried about, i.e., choose a plan $t$ such that $t \in h=\argmax_{h\in H_i} P(h )$. \\

\noindent {\bf \emph{Most Probable Plan (MPP)}.} Choose the plan that is associated with the highest cumulative  probability across all hypotheses: $\argmax_{t\in T} P(t)$, where $T$ is the  union set of  all plans in all  of the hypotheses $H$, and $P(t)$ denotes the cumulative probability assigned to all
hypotheses that contain the plan $t$, computed as follows:
\begin{equation}
P(t)=\sum_{h \in H \mid \exists p \in h, t \sim^r p} {P(h)}
\end{equation}

\noindent {\bf \emph{Minimal Entropy (ME)}.} Choose the plan with the maximal information gain (or minimal
  entropy) given  the resulting hypothesis set. The information gain directly depends on  Equations~\ref{eq:updateT}
 and~\ref{eq:updateF} for updating the hypothesis space following the results of the query $QA(p)$.
\begin{align*}
\min_{t \in T} & P(t) \cdot Ent (\phi(H_i,t,\tn{True})) +	\\
&  (1-P(t)) \cdot Ent( \phi(H_i, t, \tn{False}) )
\end{align*}
 where $Ent(\cdot)$ is the standard entropy computation over the resulting
 hypothesis space~\cite{shannon2001mathematical}.

\section{Empirical Evaluation}
\label{sec:empirical}

We evaluated the probing approaches described in the previous sections
on two separate domains from the plan recognition literature. The first is the
simulated domain used by Kabanza et
al.~\shortcite{kabanza2013controlling}. We used their same
configuration which includes 100 instances with a fixed number of
actions, five identified goals, and a branching factor of 3 for rules
in the grammar.
The second domain involves students' interactions with the {\vl} system
when solving two different types of problems: the problem
described in Section 2, and a problem which
required students to determine the concentration level of an unknown
acid solution by performing a chemical titration process. We sampled
35 logs of students' interactions in {\vl} to solve the above
problems.
%These two problems differ widely in the types of solution strategies they require from
%students, which is reflected in the length and the types of actions in
%the logs that we sampled.
In each of the logs, we used domain experts %(chemistry teachers and researchers)
to tag the  correct hypothesis.
 We used a plan-library representation
which extended basic and complex actions to include parameters, and used the
refinement methods from Amir and
Gal~\shortcite{AG13} which considered constraints over the parameter
values.
%
%with the help of domain experts (Chemistry teachers using the {\vl}
%software in their classrooms).

We used the Most Probable Plan (MPP), the Most Probable Hypothesis
(MPH) and the Minimal Entropy (Entropy) approaches, as well as a
baseline approach that picked a plan to query at random.  For both domains, we kept the PR algorithm constant as the PHATT algorithm~\cite{GeibGoldman09} and only varied the type of query mechanism used for the SPR.

We first show the number of hypotheses that were outputted by PHATT
for the various approaches, without probing interventions. As can be
seen in Table~\ref{tab:domains}, the number of hypotheses in the simulated domain grows linearly in the number of observations, but for the real-world domain,
%which is significantly more ambiguous,
the number of
hypotheses grows exponentially, reaching over 10,000 hypotheses after
just 7 actions. 

\begin{figure}[h]
 \centering
 \small
 \includegraphics[width=8cm]{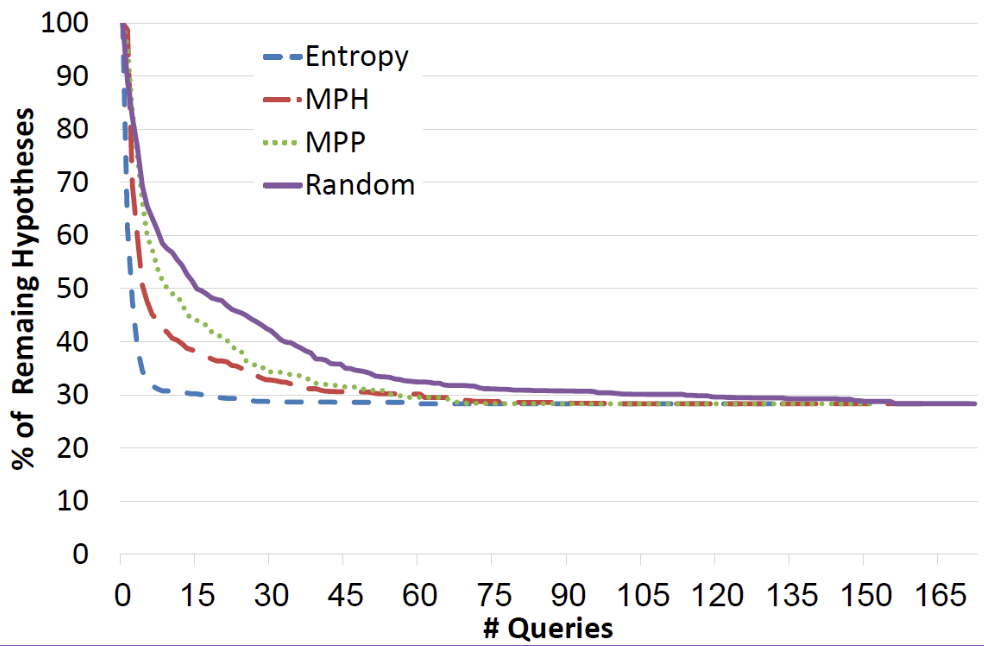}
 \includegraphics[width=8cm]{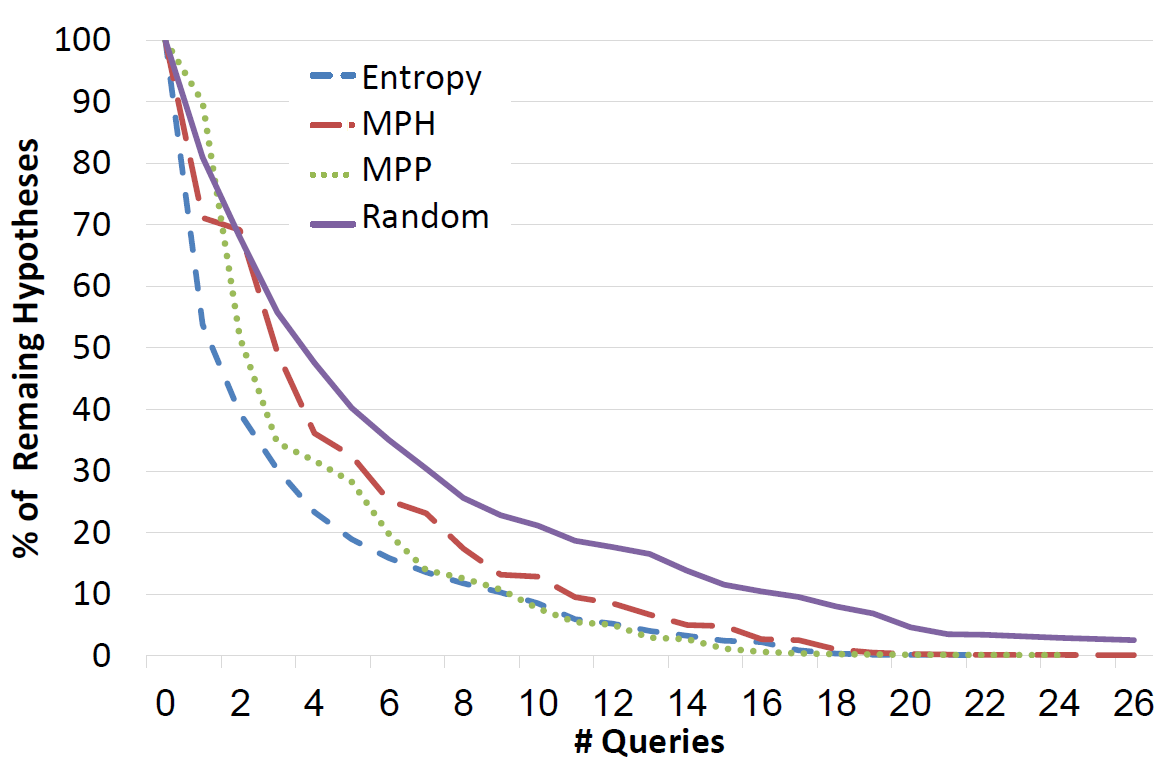}
 \caption{Decrease in the hypothesis set size after each query in the simulated domain (top) and {\vl} (bottom).
}.
 \normalsize
        \label{fig:emp}
\end{figure}

Figure~\ref{fig:emp} shows the average percentage of hypotheses remaining from the initial hypothesis set ($H_0$) as a function of the number of queries performed.  Before the first query, all algorithms start with 100\% of the  hypotheses in $H_0$, and this number decreases as more queries are performed. For both domains we used the plan recognition output after 7 observations.
\begin{table}[t]
\centering
\begin{tabular}{|l|rrrrr|}
%\footnotesize
\hline
Obs.            & 3      & 4      & 5       & 6    & 7    \\
\hline
Hyp. (VL)           & 19 & 83 & 363 & 2,011 & 11,759 \\
Hyp. (simulated)   & 12 & 25 & 28  & 32 & 25 \\
\hline
%\normalsize
\end{tabular}
\caption{Number of hypotheses per observation.}
\label{tab:domains}
\end{table}

\begin{table}
\centering
\small
\begin{tabular}{|l|rrrrr|}
\hline
Observations & 3    & 4     & 5     & 6     & 7     \\ \hline
Entropy-Sim  & *7.3 & **10.4 & **15.6 & **23.5 & **18.4 \\
MPP-Sim      & *7.3 & *10.8 & *16.3 & *25.2 & *21.0 \\
MPH-Sim      & *7.6 & *10.8 & *16.3 & *25.4 & *19.9 \\
Random-Sim   & 7.9 & 11.9 & 18.4 & 29.0 & 28.7 \\ \hline
Entropy-VL   & **7.6 & *10.7 & **13.4 & *17.2 & *18.3 \\
MPP-VL       & *8.2 & *11.2 & *14.7 & *17.7 & *18.9 \\
MPH-VL       & 8.8 & *12.2 & *16.2 & *19.8 & 21.6 \\
Random-VL    & 9.5 & 14.9 & 24.0 & 36.3 & 27.8 \\ \hline
\end{tabular}
* - significantly less queries compared to random, \\ ** - significantly less queries compared to all other strategies ($p \leq 0.05$).
\caption{Average Number of Queries until Convergence.
}
\label{tab:queries}
\end{table}

As seen in Figure~\ref{fig:emp}, both in the simulated domain and in the VL domain, the Entropy probe performed better than all other probes.
In general, all non-trivial probing techniques were able to reduce the number of hypotheses significantly compared to random, and Entropy outperformed all algorithms. As seen in the figures, although the PR process created more hypotheses for the VL domain, the convergence of SPRP is usually to a single hypothesis, while in the smaller simulated domain, all algorithms converge to a minimal hypothesis set of about 30\% of the number of hypotheses in $H_0$.  We attribute this to  inherent ambiguity in this domain that cannot be resolved by making further queries.

In general, the advantage of Entropy over all other approaches for the first five queries was statistically significant ($p\lneq0.01$). This is especially important since queries are costly and the  the number of queries that can practically be asked is  small. Thus an approach able to limit the hypotheses more with fewer queries is preferred.
Lastly, Table ~\ref{tab:queries} shows the average number of queries needed until reaching the minimal set of hypotheses, for each probing strategy. Notice that the number of hypotheses increase with each new observation. Although counter-intuitive, this is due to the fact that for each hypothesis, a new observation can initiate a new plan or complement an existing plan (or both), so the size of the hypothesis space will be at least the size of the original one.
This table shows that the  Entropy probe made significantly fewer queries than the other approaches.
\section{Conclusion}
This paper defined and studied SPRP, in which it is possible to query whether a chosen plan is part of the correct hypothesis, and subsequently remove all incorrect plans from the hypothesis space. The goal is to minimize the number of queries to converge to the minimal hypothesis set that is consistent with the observations.  We presented a number of approaches for choosing a plan to query -- the plan that maximizes the expected information gain, as well as the plan that is ranked highest in terms of likelihood by the PR algorithm. We evaluated these approaches on two domains from the literature, showing that both were able to converge to the correct hypothesis using significantly less queries than a random baseline, with the maximal information gain technique exhibiting a clear advantage over all approaches.

 We are working on extending the heuristic approach described in the paper to  using MDPs to allow for the probing policy to reason about future steps.   To this end we are working on a compact representation of a state space to represent the set of possible hypotheses. 
We also intend to use our approach to augment existing educational software to  intelligently query students about their solution strategy in a way that minimizes the disruption. We will probe each of these directions.

%\kg{we can remove the following paragraph for space, also it is just a claim that we cannot really support as we do define plan library etc... in the paper our approach does not rely on the existence of a plan library or grammar: at no point does the probing process uses the grammar or a predefined set of plans to choose its next plan to query. Rather, it only uses the outputted hypothesis set. Hence, it is naturally adaptable to other plan recognition algorithms like the generative approach suggested by Ramirez and Geffner~\cite{ramirez2009plan}, in which plans are constructed stochastically.}

% give each possible action a value, so that at each step the algorithm will
% be able to calculate the worthwhileness of querying about a plan in
%  comparison to wait for the next observation.
% Second, we assumed perfect answers to a query, which may not be
% realistic for domains which include other people.  Third, we assumed
% uniform query cost. Lastly, we only allowed asking about full
% plans. However, asking about partial plans that are not part of any
% hypothesis are also possible. We will explore all of these aspects in
% future work.  Lastly, we intend to deploy our approach in e-learning
% systems for science education, and query students about their intended
% activities.
%Our approach can be extended into various additional directions, such as MDP formalization, subplan queries and deployment in real-world settings. We are currently probing each of these directions.

\section*{Acknowledgments}
This research was funded in part by ISF grant numbers 363/12 and 1276/12, and by EU FP7 FET project no. 60085. R.M. is a recipient of the Pratt fellowship at the Ben-Gurion University of the Negev.

\bibliography{libb}
\bibliographystyle{named}

\end{document}